%% file: main.tex
\documentclass[10pt,twocolumn,letterpaper]{article}

\usepackage[pagenumbers]{cvpr} 


\definecolor{cvprblue}{rgb}{0.21,0.49,0.74}
\usepackage[pagebackref,breaklinks,colorlinks,allcolors=cvprblue]{hyperref}

\usepackage[utf8]{inputenc} 
\usepackage[T1]{fontenc}    
\usepackage{hyperref}       
\usepackage{url}            
\usepackage{booktabs}       
\usepackage{amsfonts}       
\usepackage{nicefrac}       
\usepackage{microtype}      
\usepackage{xcolor}         
\usepackage{graphicx}
\usepackage{array} 
\usepackage{colortbl}
\usepackage{bookmark}
\usepackage{wrapfig}
\usepackage{amsmath}
\usepackage{float}

\def\paperID{14921} 
\def\confName{CVPR}
\def\confYear{2025}

\title{Instruct-IPT: All-in-One Image Processing Transformer via Weight Modulation}

\author{
  Yuchuan Tian$^{1}$, Jianhong Han$^{2}$, Hanting Chen$^2$, Yuanyuan Xi$^{2}$,\\ Ning Ding$^{1}$, Chao Xu$^{1}$, Yunhe Wang$^2$\thanks{Corresponding author}\\
  \small$^1$ State Key Lab of General AI, School of Intelligence Science and Technology, Peking University. $^2$ Huawei Noah's Ark Lab.\\
  \small\texttt{tianyc@stu.pku.edu.cn, \{hanjianhong, yunhe.wang\}@huawei.com} \\ 
}

\begin{document}
\maketitle

\begin{abstract}
    Due to the unaffordable size and intensive computation costs of low-level vision models, All-in-One models that are designed to address a handful of low-level vision tasks simultaneously have been popular. However, existing All-in-One models are limited in terms of the range of tasks and performance. To overcome these limitations, we propose Instruct-IPT -- an All-in-One Image Processing Transformer (IPT) that could effectively address manifold image restoration tasks with large inter-task gaps, such as denoising, deblurring, deraining, dehazing, and desnowing. While most research propose feature adaptation methods, we reveal their failure in addressing highly distinct tasks, and suggest weight modulation that adapts weights to specific tasks. Firstly, we search for task-sensitive weights and introduce task-specific biases on top of them. Secondly, we conduct rank analysis for a good compression strategy and perform low-rank decomposition on the biases. Thirdly, we propose synchronous training that updates the task-general backbone model and the task-specific biases simultaneously. In this way, the model is instructed to learn both general and task-specific knowledge. Via our simple yet effective method that instructs the IPT to be task experts, Instruct-IPT could better cooperate between tasks with distinct characteristics at humble costs. As an additional feature, we enable Instruct-IPT to receive human prompts. We have conducted experiments on Instruct-IPT to demonstrate the effectiveness of our method on manifold tasks, and we have effectively extended our method to diffusion denoisers as well. The code is available at \url{https://github.com/huawei-noah/Pretrained-IPT}.
\end{abstract}

\section{Introduction}
The effectiveness of Transformers~\cite{transformer} has been verified on various vision tasks, including image classification~\cite{vit}, object detection~\cite{detr}, segmentation~\cite{setr}. Some works~\cite{chen2021pre,liang2021swinir} and some recent developments~\cite{zamir2022restormer,wang2022uformer,iptv2} have also introduced transformer backbones to low-level image restoration tasks. Although transformers proposed in these works are powerful low-level vision models, they are only experts on one single task. In real applications, however, several tasks ought to be addressed by the same system, and it is overly tedious to replicate several heavy low-level vision transformer models for different tasks.

To address several different tasks, previous works have proposed All-in-One image restoration models, \textit{i.e.} models where several tasks share the same backbone. All-in-One works like AirNet~\cite{li2022all} and PromptIR~\cite{potlapalli2023promptir} are indeed smarter in the sense that their backbones are universally applicable to three canonical low-level tasks, but they are suffering limitations as well. Firstly, the scope of their application is limited. All-in-One models are confined to only a handful of conventional low-level tasks, leaving other restoration tasks behind. Secondly, their performance of each individual task is limited. In spite of task-adapting strategies, the tasks are impeding with each other within a shared model backbone. 

\begin{figure*}[!t]
    \centering
    \includegraphics[width=\textwidth]{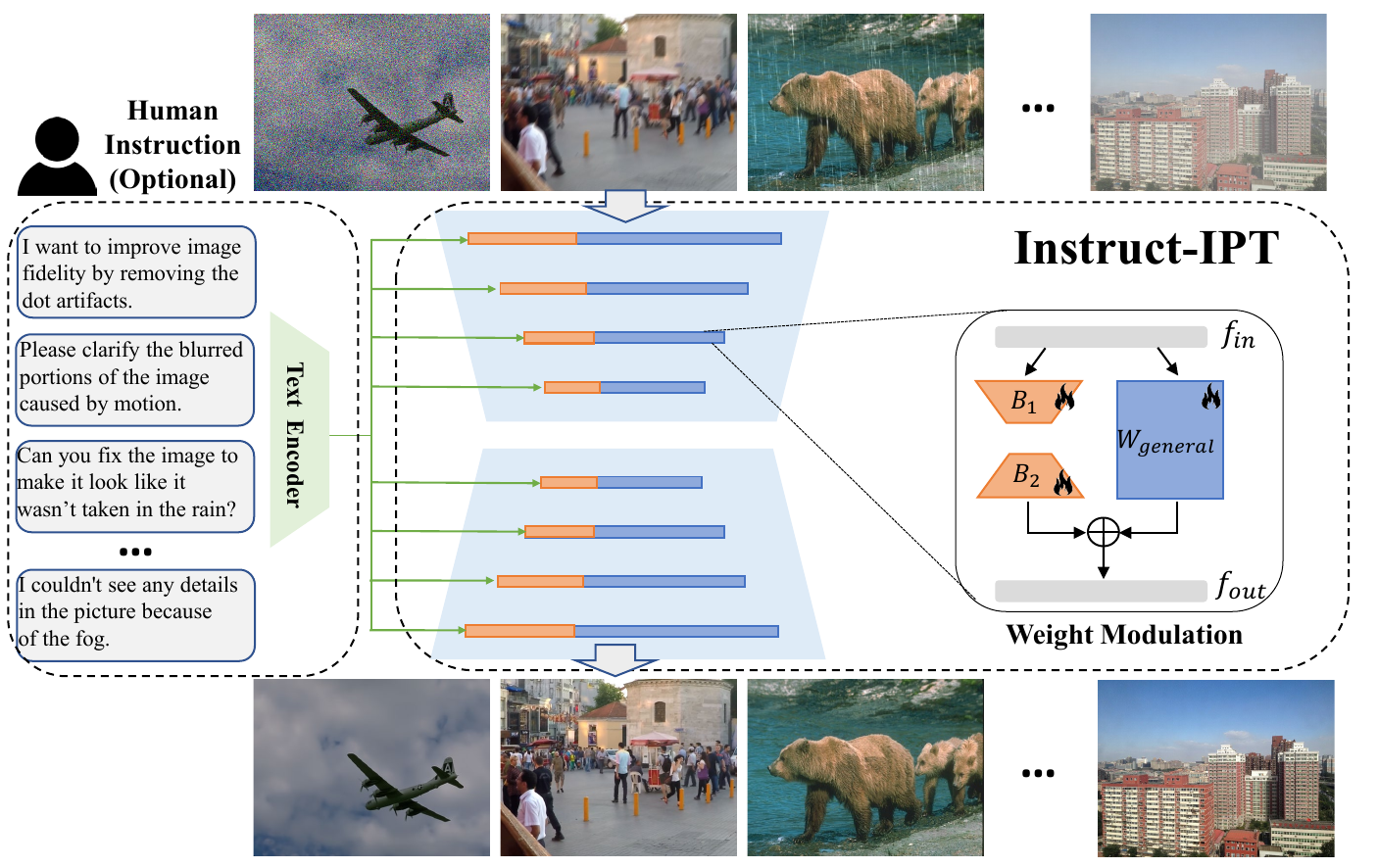}
    \caption{\textbf{Framework of Instruct-IPT}. Thanks to the proposed weight modulation method, Instruct-IPT performs well on a wide range of tasks. Weight modulation involves adding task-specific biases (which is low-rank decomposited) to a general backbone. Synchronous training is performed where both the backbone and the bias are updated simultaneously, such that task-specific knowledge is automatically extracted. Text instructions could be provided to command the model.}
    \label{fig:main}
  \end{figure*}
  
  The deep-rooted reason behind these limitations is the ineffectiveness of existing adaptation methods for different low-level vision tasks. Most existing methods could address tasks that are highly related to each other, but they fail for tasks that are different in nature. As a result, including a task that is principally different could pull the overall low-level task performance down. The application value of previous All-in-One models is thus limited.
  
  To overcome these limitations, we tend to renovate previous All-in-One architectures for more tasks and better adaptation to specific tasks. To this end, we propose \textbf{Instruct-IPT}, an All-in-One solution with the Image Processing Transformer (IPT) that effectively addresses a wide range of image restoration tasks. First and foremost, we conclude from experiments that weight rather than feature modulation to specific tasks is the right choice for tasks that are highly irrelevant. Then we propose a simple yet effective weight modulation method, \textit{i.e.} the addition of a task-specific bias to the weight backbone. For feasibility, low-rank decomposition is proposed on the basis of a rank analysis. We also propose synchronous training, which naturally extracts single-task knowledge from general knowledge. The weight modulation is introduced to the Image Processing Transformer, on top of which we introduce text instructions to make Instruct-IPT multimodal. Instruct-IPT is capable of responding to human instructions and is thus more adaptable to real-world scenarios. We have conducted extensive experiments to verify the performance of Instruct-IPT. We further extend our method to diffusion models in experiments, demonstrating the wide applicability of our method.

  \section{Related Work~\label{sec:related}}
  \textbf{Transformer-based Image Restoration Methods.} Transformers have pushed image restoration performance further beyond canonical CNN counterparts. Several transformer-based approaches~\cite{chen2021pre,li2022all,li2020all,park2023all,wang2022uformer} have been proposed to tackle various degradation restoration tasks. Among these methods, IPT~\cite{chen2021pre} first introduced a standard transformer-based framework with multi-heads and multi-tails structure to handle multiple image restoration tasks. SwinIR~\cite{liang2021swinir} employed Swin-transformer blocks with a shifted window scheme tailored for restoration tasks, while Uformer~\cite{wang2022uformer} combined U-Net and transformer with window-based self-attention to improve model performances. Other transformer-based methods, such as Restormer~\cite{zamir2022restormer} and IPT-V2~\cite{iptv2}, leveraged channel self-attention to capture longer-range dependencies. 
  
  \noindent\textbf{All-in-One Image Restoration Methods.} Manifold existing works have explored the restoration of images corrupted by multiple degradations. These works could be divided into two classes based on their approach: feature adaptation and weight modulation. Feature-based measures that receive popularity usually tailor intermediate features for a certain task using task-specific contextual information (like prompts). Among these methods~\cite{li2022all,gao2024prompt,kong2024towards,li2023prompt,ma2023prores,potlapalli2023promptir}, contextual information might include images, degradation context, and input features. The other class of weight modulation measures tailor network weights for certain tasks. Park et al.~\cite{park2023all} proposed ADMS, which employed adaptive unstructured-sparse filters with independent parameters for different degradations. Zhu et al.~\cite{zhu2023learning} propose using different sets of neurons for different tasks. A shared drawback of all these measures is that they are applied to a handful of tasks (usually three) that are highly relevant, and their performance is not comparable to single-task image restoration models. Hence, we are investigating a method that could be applied to distinct tasks with outstanding performance.

  \noindent\textbf{Text Commands in Image Restoration.} Text commands were traditionally used as a semantic instructor to models. Text-to-image generation that generates images based on the semantic meaning of texts has been widely studied in the realm of diffusion models~\cite{ramesh2022hierarchical,rombach2022high,saharia2022photorealistic,zhang2023adding}. Most of these models are pre-trained on diverse text and image data pairs, resulting in an enhanced understanding of textual information. As diffusion models have also been applied to low-level vision tasks for image restoration~\cite{brooks2023instructpix2pix,wang2023exploiting,wu2023seesr}, text commands have been used to guide the model's judgment on the type of image degradation beyond facilitating image content generation. For instance, PromptSR~\cite{chen2023image} introduced additional text priors that encoded the way of degradation; Lin~\cite{lin2023improving} proposed an SD-based image restoration method that introduces text-based degradation priors in addition to image-based content priors for multi-task image restoration. TIP~\cite{qi2023tip} employed degradation-related text instructions as well. Apart from the application of text commands on diffusion, InstructIR~\cite{conde2024high} also introduces text instructions to conventional image restoration models. Endorsing the advantage of human language as a better user interface, we follow this path and add the feature of text instructions to our Instruct-IPT.

  \section{Method}
  
  In this section, we go through the details of the proposed Instruct-IPT method that adapts the image processing transformer to various different tasks. First and foremost, we are faced with the choices of feature-based or weight-based selection methods. 
  
  \subsection{Evaluating Feature Adaptation Methods.}
  
  As is introduced in Sec.~\ref{sec:related}, adapting feature maps to certain tasks for multi-task image processing models has been widely discussed. However, the tasks discussed in these works are usually highly related (e.g. denoising-deraining-dehazing). On the other hand, since we are interested in a model that could handle a broader range of tasks, how would these feature adaptation methods work on low-level tasks that are highly irrelevant? 
  
  To this end, we experiment with these feature adaptation methods by selecting two distinct image restoration tasks: image denoising and motion deblurring. While image denoising aims at removing random noises on the image, motion deblurring overcomes the blurring effect from a motion blur kernel. Judging from their visual effect and principles, we tend to consider them as highly distinct. Hence, we observe the performance of previous adaptation methods on these two tasks.
  
  We have selected a pool of the latest multi-low-level-task adaptation methods as follows. \textbf{Prompting} from PromptIR~\cite{potlapalli2023promptir}: task-specific prompts are injected at each U-Net stage transition; \textbf{Instruct-Channel} from InstructIR~\cite{conde2024high}: a series of task-specific weights perform channel-wise affine feature mapping; \textbf{External-Control} from ControlNet~\cite{zhang2023adding}: a degradation-sensitive parallel encoder network is added that intervenes features in the original backbone. Notably, these adaptation methods all modify intermediate features for certain tasks. We align these feature-based methods to the IPT-V2~\cite{iptv2} model. As an improved version of IPT~\cite{chen2021pre}, IPT-V2~\cite{iptv2} achieves outstanding performance on low-level image restoration tasks, and thus being a good base model for our experiment.

  \begin{table}[htbp]
    \centering
    \setlength{\belowcaptionskip}{0cm}   
    \begin{tabular}{lrr}
      \toprule
      Layers & Similarity \\\midrule
      \textbf{Image Processing} &  \\
      \quad Image Embedder & \textcolor{blue}{0.97}  \\
      \quad Output Layer & \textcolor{red}{0.68}  \\\midrule
      \textbf{U-Net Layers} &  \\
      \quad UpSampling \& DownSampling & \textcolor{red}{0.52}  \\
      \quad Channel Reduction & 0.73  \\\midrule
      \textbf{Transformer Layers} &  \\
      \quad Query-Key-Value Projection & 0.72  \\
      \quad Post-Attention Projection & 0.80  \\
      \quad Feed-Forward-Network Projection & 0.73  \\
      \quad LayerNorm & \textcolor{blue}{0.99}  \\
      \bottomrule
    \end{tabular}%
    \vspace{-5pt}
    \caption{\textbf{Similarities of weights pre or post finetuning.} \textcolor{red}{Red}/\textcolor{blue}{Blue} weights are \textcolor{red}{more}/\textcolor{blue}{less} sensitive to specific tasks.}
    \label{tab:tasksim}%
  \end{table}

  \begin{table*}[htbp]
    \centering
    \setlength{\belowcaptionskip}{0cm}   
    \begin{tabular}{lcccc}
      \toprule
      & \multicolumn{2}{c}{Denoising} & \multicolumn{2}{c}{Deblurring} \\
      Methods & BSD~\cite{martin2001database} & Urban100~\cite{huang2015single} & GoPro~\cite{nah2017deep} & HIDE~\cite{shen2019human} \\
      \midrule
      Plain Mixed Training & 34.37 & 35.12 & 32.81 & 30.75 \\
      \textbf{Prompting} from PromptIR~\cite{potlapalli2023promptir} & 34.37 & 35.13 & 32.85 & 30.81 \\
      \textbf{Instruct-Channel} from InstructIR~\cite{conde2024high} & 34.31 & 34.93 & 33.11 & 31.06 \\
      \textbf{External-Control} from ControlNet~\cite{zhang2023adding} & 34.38 & 35.14 & 32.64 & 30.57 \\\midrule
      Instruct-IPT (Ours) & \textbf{34.40} & \textbf{35.19} & \textbf{33.86} & \textbf{31.65}  \\
      \bottomrule
    \end{tabular}
    \vspace{3pt}
    \caption{\textbf{Experimenting feature adaptation methods on distinct tasks.} We mix two tasks for training as the baseline, and compare a series of recent feature adaptation methods against it. For fair comparison, all methods are applied to the same backbone and trained for the same \# of iterations.}
    \label{tab:feature-based}
  \end{table*}
  
  The results of this experiment turn out to be appalling: while these All-in-One adaptation methods claim to be effective on multiple tasks (e.g. denoising, deraining, \& dehazing), they are not performing well when the task of denoising and deblurring is combined. Specifically, most methods struggle in the balance between two tasks: they usually sacrifice one task for the other. For instance, InstructIR sacrifices denoising for deblurring, and External-Control sacrifices deblurring for denoising. 
  
  Existing feature-based adaptation methods are insufficient in tailoring the model for tasks that are highly different. As the performance of feature adaptation methods is unsatisfactory, we opt for weight adaptation that directly modifies weights.
  
  \subsection{Efficient Weight Adaptation\label{sec:efficient_weight_adaptation}}

  \textbf{Picking Task-Sensitive Weights.}
  Previously, there are some methods that modifies weights for specific tasks~\cite{park2023all,zhu2023learning}. However, these methods universally apply weight modification on all types of weights without evaluating their individual contribution to different tasks. 
  
  To perform the evaluation, we finetune the weights on a specific task and then compare the average cosine similarity of weights after and before finetuning. The weights in IPT could be classified into three:

  1. \underline{Image processing components:} This part contains the basic elements of an image processing model, including an image embedder and the last output layer that maps high-dimensional features to restored images.
  
  2. \underline{U-Net layers:} This group contains key components in the U-Net architecture, including upsampling, downsampling, and channel reduction convolution (after concatenation of backbone features with shortcuts).
  
  3. \underline{Transformer layers:} This group contains key components in the transformer architecture (IPTBlock), including Query-Key-Value tuple mapping, Feed-Forward-Network (FFN) mapping, normalization layers, and after-attention fully-connected projection.
  
  Results in Tab.~\ref{tab:tasksim} turn out that Up\&Downsampling convolutions and the final output layer are highly task-sensitive. On the other hand, the image embedder for deep features and layer-norm weights are not modified too much by task finetuning. Hence, we omit these weights and modulate other weights for specific tasks.

  \noindent\textbf{The Proposed Weight Modulation.} The simple idea behind weight modulation is to add task-specific biases on top of IPT weights to adapt them to various tasks. For instance, for a weight $W'$ that performs linear mapping on a certain task, we decompose it into the matrix addition of the weight for general restoration task knowledge $W_\text{general}$ and task-specific modulation bias $B_\text{task}$ as follows:
  
  \begin{equation}
    \label{equ1}
  Y = W'X = (W_\text{general} + B_\text{task})X, 
  \end{equation}
  
  However, the naive addition of biases is impractical due to heavy parameters. Among all types of weights, fully connected layers in FFNs are particularly parameter rich, accounting for 50\% of the overall parameters. Hence, we resort to a parameter-efficient measure to modify weights at similar costs. 
  
  Previous measures~\cite{park2023all} used unstructured-sparse biases for compression, but this involves the training of dense biases that bring huge additional costs. Inspired by previous low-rank decomposition works~\cite{rank5_13,rank35_zhangxiangyu,lora}, we resort to structured compression of biases by decomposing it into the multiplication of two low-rank matrices: for $B_{task} \in \mathbb{R}^{n_\alpha \times n_\beta} $, given low-rank $n_\gamma \ll \min(n_\alpha, n_\beta)$, we have
  
  \begin{equation}
  \label{equ2}
  B_{1} \in \mathbb{R}^{n_\alpha \times n_\gamma} \land B_{2} \in \mathbb{R}^{n_\gamma \times n_\beta} \quad \text{s.t.}\quad B_\text{task} = B_1 B_2. 
  \end{equation}
  
   To demonstrate the feasibility of rank-based compression and figure out an effective measure for rank selection, we conduct rank analysis on $B_{task}$ when the pretrained model is finetuned on a specific task without rank constraints.
  
   \noindent\textbf{Rank Analysis.} In the previous toy experiment, we analyze weight similarities when the same pretrained model is finetuned on different tasks without constraints. To inspect the change of weights for task-specific purposes, we perform rank analysis on the difference between the finetuned weights and the pretrained weights (\textit{i.e.} task-specific biases in the last paragraph). Due to varying weight shapes in different U-Net stages, we are interested in the energy distribution on the spectrum. We select a rank $r$ for a weight $W$, and calculate its PCA accumulative energy $E$ of weights (normalized by Frobenius Norm) on that rank as follows: 
  
  \begin{equation}
    \label{equ3}
    \begin{aligned}
    E = \sum^{i=1}_{r} \sigma_i^2 \quad
    \text{s.t.}&\quad \frac{W}{\|W\|_F} = \sum^{i=1}_{k} u_i^T\sigma_i v_i,\\\text{where}&\quad U^T U = I, V^T V = I.
    \end{aligned}
  \end{equation}
  
  \begin{figure*}[htbp]
    \centering
    \includegraphics[width=\textwidth]{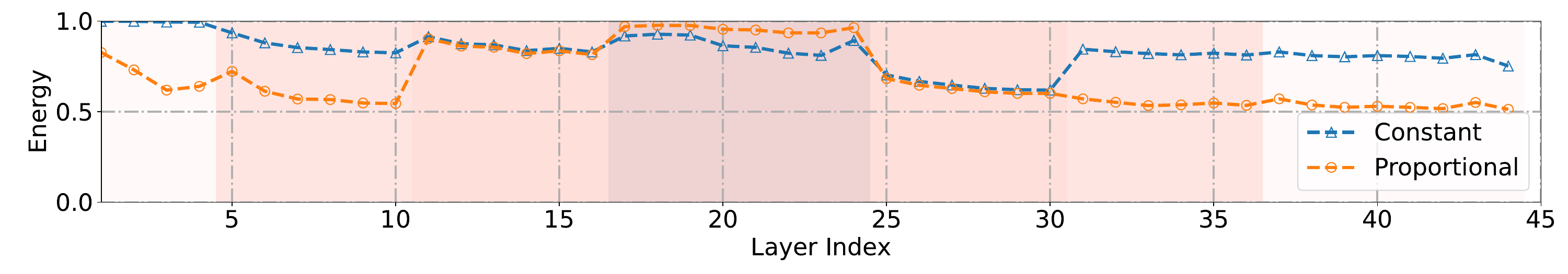}
    \vspace{-20pt}
    \caption{\textbf{PCA accumulative energy under different rank strategies across layers.} The shade of the background color indicates the depth of the U-Net stage. The constant rank strategy is better than proportional strategy in covering the overall information of biases.}
    \label{fig:PCA}
  \end{figure*}

  \begin{table*}[htbp]
    \centering
    \setlength{\belowcaptionskip}{0cm}   
    \begin{tabular}{lcccc}
      \toprule
      & \multicolumn{2}{c}{Denoising} & \multicolumn{2}{c}{Deblurring} \\
      Methods & BSD~\cite{martin2001database} & Urban100~\cite{huang2015single} & GoPro~\cite{nah2017deep} & HIDE~\cite{shen2019human} \\
      \midrule
      Plain Mixed Training & 34.37 & 35.11 & 32.81 & 30.75 \\
      \qquad + Ft. Bias (Denoising) & 34.38 & 35.15 & - & - \\
      \qquad + Ft. Bias (Deblurring) & - & - & 33.40 & 31.16 \\
      \midrule
      Sync. Training & \textbf{34.40} & \textbf{35.19} & \textbf{33.86} & \textbf{31.65}  \\
      \bottomrule
    \end{tabular}
    \vspace{-5pt}
    \caption{\textbf{Comparing Synchronous Training with conventional Two-Stage Training (Training \& Finetuning).} The overall training iterations is the same for all baselines. The results demonstrate that Synchronous Training, \textit{i.e.} training both weight and bias simultaneously, performs better than the Two-Stage process.}
    \label{tab:synctraining}
  \end{table*}

  The PCA accumulative energy of a matrix is an indicator of its rank: the convergence of energy to 1 at a certain rank means the matrix could be low-rank approximated at that rank. Here we provide two options for rank selection on U-Net: 1. Constant Rank: We select a constant as rank $r$ regardless of the varying weight shapes in U-Net; 2. Proportional Rank: we select a proportion $p$ between 0 and 1, such that given the rank of a weight as $k$, we evaluate the PCA accumulative energy of $r=\operatorname{Round}\left(p k\right)$. We plot their PCA accumulative energy, as shown in Fig.~\ref{fig:PCA}. From the plot, it can be revealed that the constant-rank strategy is better. Proportional rank selection is not performing well on shallow layers: the chosen rank could result in 30\% to 50\% of the information loss. Constant rank selection performs uniformly well in covering the overall energy of task-specific biases.

  \noindent\textbf{Synchronous Training.}
  In previous methods, task-specific modification learning is usually conducted on top of a pretrained model for various tasks. This widely-applied training process forcibly separates task-specific knowledge learning from general knowledge learning, featuring a two-stage process. However, these methods might not perform well, because low-level vision models cannot be effectively pretrained due to large inter-task gaps. Rather than the redundant two-stage solution, we hope the model could learn task-specific and general knowledge at the same time.
  
  As a simple yet effective measure, we unfreeze the backbone and train both the backbone and the task-specific biases synchronously. We conduct synchronous training on IPT-V2~\cite{iptv2} and compare it with the conventional two-stage method. Experiment results in Tab.~\ref{tab:synctraining} prove the effectiveness of our method.

  \begin{table*}[htbp]
    \centering
    \setlength{\belowcaptionskip}{0cm}   
    \setlength\tabcolsep{4pt}
  \begin{tabular}{lccccccc}
    \toprule
    Task & \multicolumn{3}{c}{Denoising (BSD68)~\cite{martin2001database})} & Deblurring & Deraining & Dehazing & Desnowing \\
    Model & $\sigma=15$ & $\sigma=25$ & $\sigma=50$ & GoPro~\cite{nah2017deep} & Rain100L~\cite{yang2020learning} & SOTS~\cite{li2018benchmarking} & CSD~\cite{chen2021all} \\
    \midrule
    \textbf{AirNet} & 33.92 & 31.26 & 28.00 & - & 34.90 & 27.94 & -\\
    \textbf{NAF-ADMS-1} & - & \underline{31.53} & - & 29.99 & 33.15 & - & - \\
    \textbf{NAF-ADMS-2} & - & - & - & - & 31.89 & 30.56 & \underline{33.83} \\
    \textbf{CAPTNet} & - & 30.75 & - & \underline{32.71} & \underline{37.86} & 29.28 & - \\
    \textbf{PromptIR} & \underline{33.98} & 31.31 & \underline{28.06} & - & 36.37 & \underline{30.58} & - \\
    \textbf{InstructIR} & - & 31.09 & - & 26.65 & 35.58 & 25.20 & - \\
    \textbf{InstructIR-5D} & - & 31.40 & - & 29.40 & 36.84 & 27.10 & - \\
    \textbf{Instr.-IPT-Tiny (Ours)} & {34.27} & {31.66} & {28.48} & {32.00} & {37.09} & {37.51} & {36.66} \\
    \textbf{Instr.-IPT (Ours)} & \textbf{34.40} & \textbf{31.79} & \textbf{28.61} & \textbf{33.86} & \textbf{37.88} & \textbf{39.95} & \textbf{40.12} \\
    \bottomrule
    \end{tabular}
    \vspace{-5pt}
    \caption{\textbf{Comparing Instruct-IPT with All-in-One models on 5 low-level vision tasks.} We select a bunch of the latest All-in-One methods and evaluate them on five task benchmarks, including denoising, deblurring, deraining, dehazing, and desnowing. The best and second best results of multi-task restoration are \textbf{bolded} and \underline{underlined}, respectively.}
    \label{tab:all-in-one}
  \end{table*}
  
  \section{Experiments\label{sec:experiments}}
  
  In this section, we conduct experiments to demonstrate the outstanding performance of the proposed Instruct-IPT model.

  \noindent\textbf{Implementation details:} Our training set is a large combination of various datasets from multiple tasks, mainly following the datasets in All-in-One image restoration methods~\cite{potlapalli2023promptir,park2023all,chen2021pre}. Our training hyperparameters largely follow IPT-V2~\cite{iptv2}. We classify all tasks into two parts: elementary and downstream tasks. We hold that denoising and deblurring are elementary tasks that contribute to other downstream tasks. Hence, we perform synchronous training on these two tasks to learn weights for general knowledge and biases for specific knowledge. Accordingly, the training is lengthened by a factor of two. We simply finetune biases for other tasks. We use 8 NVIDIA A100 GPUs.
  
  As for text commands, we employed a Large Language Model (LLM) to generate a substantial dataset of text instructions. Subsequently, human annotators filtered out items with ambiguous meanings, resulting in a final corpus of 2,000 text instructions per task.
  
  \begin{table}[H]
    \centering
    \setlength{\belowcaptionskip}{0cm}   
  \begin{tabular}{lccc}
    \toprule
    Model & FLOPs & Latency & Params \\
    \midrule
    \textbf{AirNet} & 19.47G & 0.0373s & 8.93M \\
    \textbf{InstructIR} & 1.02G & 0.0259s & 15.84M \\
    \textbf{PromptIR} & 10.83G & 0.0751s & 35.59M \\
    \textbf{Ins.IPT-Tiny (Ours)} & 0.95G & 0.0193s & 2.60M \\
    \textbf{Ins.IPT (Ours)} & 10.01G & 0.0598s & 26.49M \\
    \bottomrule
  \end{tabular}
    \vspace{-5pt}
    \caption{\textbf{FLOPs, latency, and \# of parameter statistics of baselines.} In accordance to baselines, we design two versions of Instruct-IPT to match their sizes.}
    \vspace{-10pt}
    \label{table:model}
  \end{table}

  \begin{table*}[htbp]
    \centering
    \setlength{\belowcaptionskip}{0cm}   
  
    \begin{tabular}{c c c c c c c | c}
      \toprule
      RESCAN & PreNet & MPRNet & MSPFN & SPAIR & Restormer & M3SNet & Instruct-IPT \\
      \cite{li2018recurrent} & \cite{ren2019progressive} & \cite{zamir2021multi} & \cite{jiang2020multi} & \cite{purohit2021spatially} & \cite{zamir2022restormer} & \cite{m3snet} & (Ours) \\
      \toprule
      29.80 & 32.44 & 32.40 & 36.40 & 36.93 & 38.99 & \textbf{40.04} & \underline{39.35} \\
      \bottomrule
    \end{tabular}
  
    \vspace{0.2cm} 
    
    \scalebox{0.8}{
      \begin{tabular}{c c c c c c c c c | c}
      \toprule
      AOD-Net & Uformer & GridDehazeNet & FFA-Net & MAXIM & DehazeFormer & IRNeXt & SFNet & FSNet & Instruct-IPT \\
      \cite{li2017aod} & \cite{wang2022uformer} & \cite{liu2019griddehazenet} & \cite{qin2020ffa} & \cite{tu2022maxim} & \cite{song2023vision} & \cite{cui2023irnext} & \cite{cui2022selective} & \cite{cui2023image} & (Ours) \\
      \toprule
      24.14 & 26.52 & 30.86 & 33.57 & 34.19 & 34.95 & 39.18 & \underline{40.05} & \textbf{40.40} & 39.95 \\
      \bottomrule
    \end{tabular}}
  
    \vspace{0.2cm} 
    
    \begin{tabular}{c c c c c c | c}
      \toprule
      DesnowNet & HDCW-Net & Uformer & Restormer & NAFNet & SnowFormer & Instruct-IPT \\
      \cite{desnownet} & \cite{chen2021all} & \cite{wang2022uformer} & \cite{zamir2022restormer} & \cite{chen2022simple} & \cite{snowformer} & (Ours) \\
      \toprule
      20.13 & 29.06 & 33.80 & 35.43 & 35.13 & \underline{39.45} & \textbf{40.12} \\
      \bottomrule
    \end{tabular}
    \vspace{-5pt} 
    \caption{\textbf{Comparing Instruct-IPT with downstream task experts. Upper: Deraining comparison.} PSNR (on Y) on the Rain100L~\cite{yang2020learning} is reported. \textbf{Middle: Dehazing comparison.} PSNR on the SOTS Outdoor dataset~\cite{li2018benchmarking} is reported. \textbf{Lower: Desnowing comparison.} PSNR on the CSD~\cite{chen2021all} desnow dataset is reported. The best and second best results are \textbf{bolded} and \underline{underlined}.}
    \label{tab:singleexpert}
  \end{table*}

  \noindent\textbf{Benchmarks and evaluation metric:}
  For the image denoising task, we conduct testing on the BSD68~\cite{martin2001database} dataset. We generate noisy images by adding Gaussian noise  to clean images with different noise levels $\sigma \in \{15, 25, 50\}$. In the image  deblurring task, we utilize the well-known GoPro~\cite{nah2017deep} dataset, which consists of 1111 images for testing.
  For the image deraining task, we employ the Rain100L~\cite{yang2020learning} dataset, which contains 100 pairs of original images and their corresponding rainy images. 
  In the image dehazing task, we utilize the standard outdoor test set of the SOTS~\cite{li2018benchmarking} dataset, which consists of 500 images for testing.
  Finally, for the image desnowing task, we use the CSD~\cite{chen2021all} dataset as a benchmark. 
  For all tasks in the following experiments, we employ PSNR as the universal evaluation metric to comprehensively assess the effectiveness of the algorithms.

  \noindent\textbf{Overhead Comparison.} We report the FLOPs, latency, and parameter information of mainstream baselines in Tab.~\ref{table:model}. As all-in-one works seldom report overhead statistics, most statistics are measured by us from their opensourced codes. To match the models, we propose two versions of Instruct-IPT: the tiny version matches the powerful baseline of InstructIR~\cite{conde2024high}, while the normal version is aligned to PromptIR~\cite{potlapalli2023promptir}.

  \subsection{Comparison with All-in-One Methods}
  Tab.~\ref{tab:all-in-one} presents a comparative analysis of various image restoration models across five different tasks, including image denoising, deblurring, deraining, dehazing, and desnowing. The table reports PSNR for each model on different benchmarks. Among the evaluated models, our Instruct-IPT distinguishes itself by its remarkable efficacy and versatility. It achieves the highest performance across all five tasks. Our method achieves the highest PSNR values under all three noise levels, especially under high noise levels ($\sigma=50$). Our method outperforms the second-best results by a large margin of 1.15 dB, 3.18 dB, and 9.37 dB in the tasks of deblurring, desnowing, and dehazing, respectively. Overall, the strong performance of our method across denoising, deblurring, deraining, dehazing, and desnowing tasks demonstrates its high generalization capability and robustness. It can effectively handle various image degradation problems and produce high-quality restored images.

  \begin{figure*}[!t]
    \centering
    \includegraphics[width=1\textwidth]{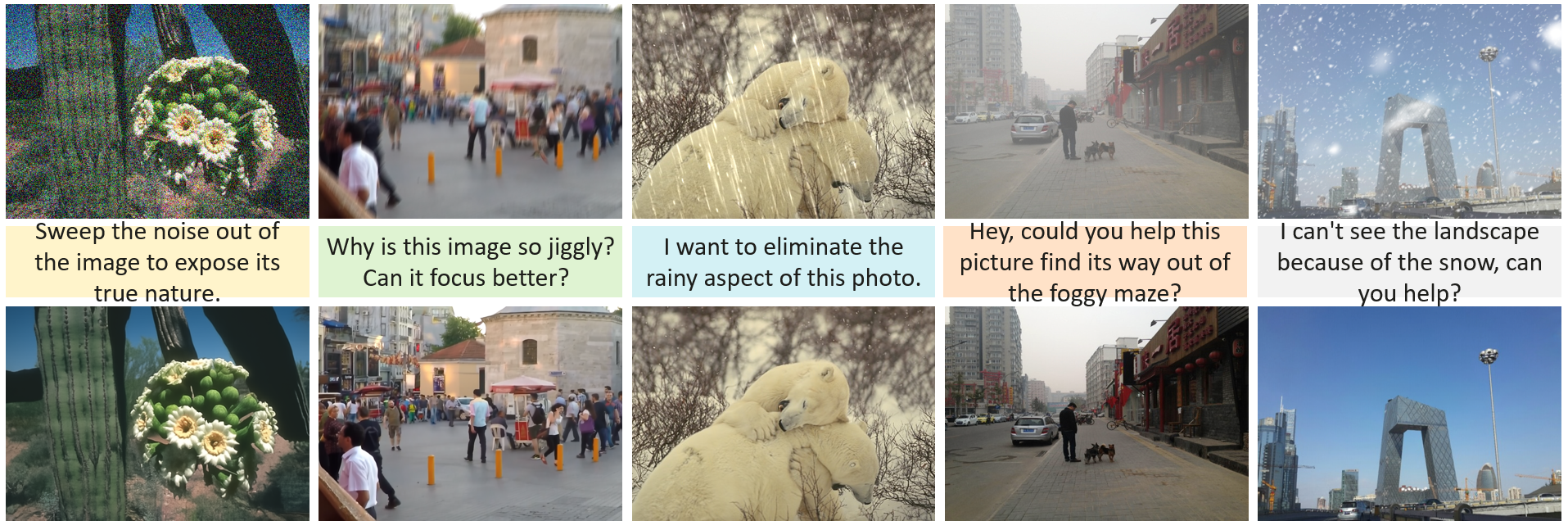}
    \caption{\textbf{A demo of Instruct-IPT instructed by human language.} Our method could achieve good image restoration results on various tasks while responding to human language. }
    \label{fig:textinteraction}
\end{figure*}

  The advanced performance stems from its innovative training strategy, which combines a general weight shared across all tasks with task-specific biases. This approach effectively facilitates the learning of task-relevant information while mitigating the detrimental impact of the large gap between different tasks. Additionally, its single-stage training paradigm enhances efficiency, reducing training time and resource consumption compared to traditional two-stage methods. Its versatility makes Instruct-IPT a promising candidate for real-world applications, particularly in scenarios demanding efficient and effective solutions for diverse image restoration needs.

  \subsection{Comparison with Single-Task Experts}
  
  Tab.~\ref{tab:singleexpert} presents a series of comparisons of our proposed Instruct-IPT, trained in an All-in-One manner, against a suite of expert models specifically trained for each individual task. Our results demonstrate that Instruct-IPT achieves competitive performance, approaching the state-of-the-art results of expert models. While not yet surpassing the current best performance, Instruct-IPT ranks among the top-tier techniques, underscoring its efficacy and potential for further development. This analysis highlights the viability of our unified training approach as a compelling alternative to task-specific models, particularly when considering the balance between model generalizability and task-specific performance. Due to page limits, some tables are presented in the Appendix.
  
  \subsection{Human Instruction}
  Further beyond weight modulation, we inject natural language as commands into the model. In reality, images are usually judged by humans in a subjective manner: the demands of humans might vary under different circumstances. Thus we hope language commands, as a better user interface, could enable complicated demands in an interactive manner. Thanks to the development of Large Language Models (LLMs), we could leverage their strong capabilities to generate a large number of authentic human commands. Then we are able to finetune the additional text encoder with these commands, which responds well to human language. As shown in Fig. ~\ref{fig:textinteraction}, the proposed model could respond to various casual instructions from humans. Sometimes, human language can be fairly casual, but Instruct-IPT is still able to respond accurately based on human instructions. This feature enables real-world applications in our daily life.

\begin{figure*}[htbp]
  \centering
  \includegraphics[width=1\textwidth]{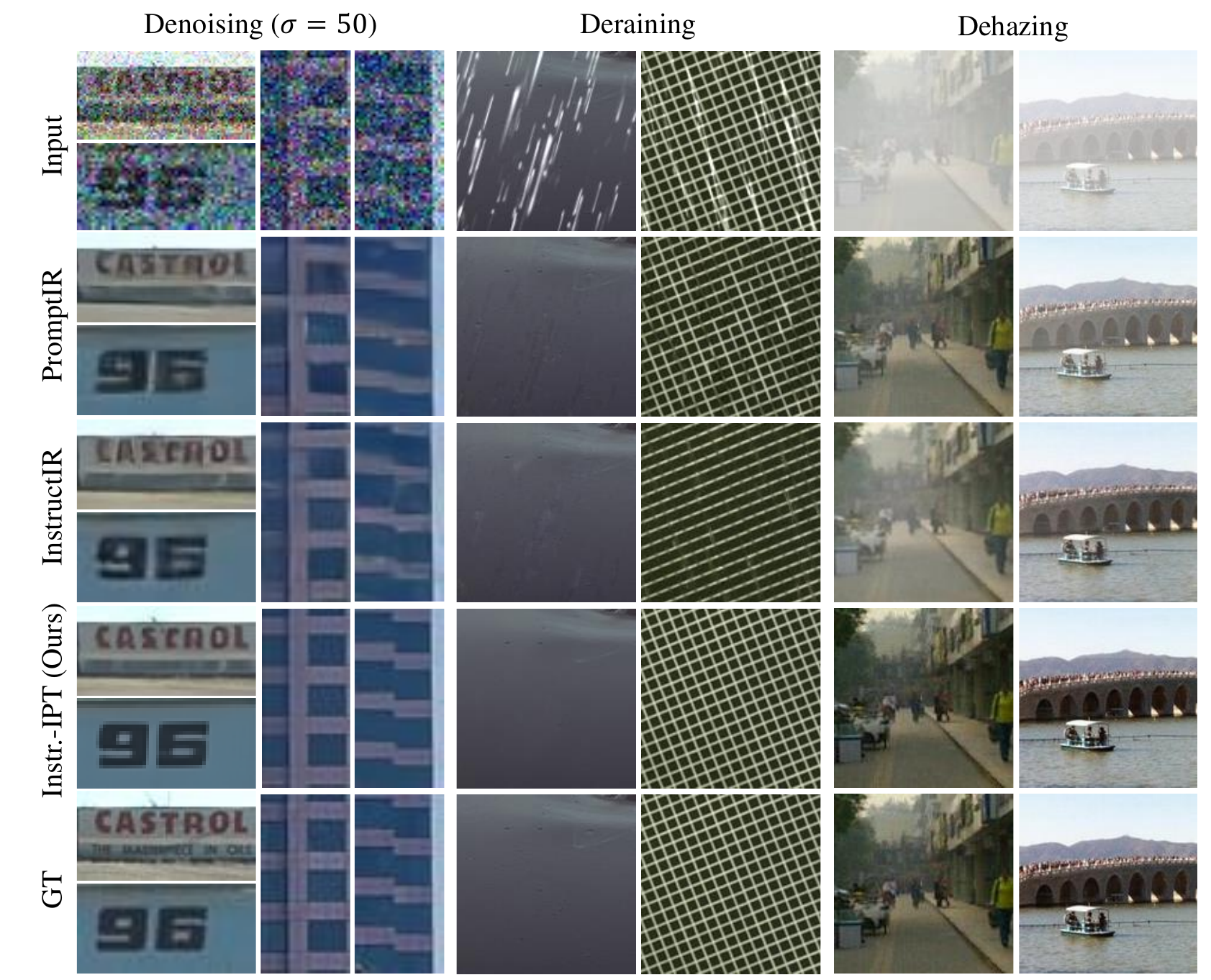}
  \vspace{-5pt}
  \caption{\textbf{Qualitative comparisons between Instruct-IPT and competitive baselines.} We compare the two methods on three tasks: denoising ($\sigma=50$), deraining, and denoising. Our Instruct-IPT could outcompete baselines by large margins in terms of visual quality.}
  \label{fig:ipt}
\end{figure*}

  \subsection{Extrapolation to Diffusion Models}
  The method we developed for IPT can be extrapolated to the U-Net architecture within diffusion models, demonstrating its generalizability. We leverage UniControl~\cite{qin2023unicontrol} as a baseline for validating our approach within the diffusion model setting,  UniControl is a multi-task unified framework based on Stable Diffusion, capable of handling diverse image generation tasks. For our evaluation, we selected inpainting and outpainting as representative tasks. Due to the absence of standardized benchmarks for these tasks, we employ COCO-Stuff~\cite{caesar2018coco} as our test dataset. 
  
  We compared the performance of our proposed method against several baseline approaches: inference with original weights, direct fine-tuning, and fine-tuning only task-specific weight biases. We adopt metrics including LPIPS (Learned Perceptual Image Patch Similarity)~\cite{zhang2018perceptual}, DISTS (Deep Image Structure and Texture Similarity)~\cite{Ding_2020}, FID (Fréchet Inception Distance)~\cite{heusel2018gans}, and IS (Inception Score)~\cite{salimans2016improved}. Our results demonstrate that our method consistently outperforms these baselines across all evaluated metrics. Due to page limits, we put the table in Appendix. We also present a visual qualitative comparison in the appendix, showcasing our method's ability to produce more visually compelling and appealing results compared to other baselines.



\section{Conclusion}
  Existing All-in-One image restoration methods are limited both in terms of task scope and performance. In this paper, we propose an All-in-One method that has outstanding performance on a wide range of image restoration tasks. We start by benchmarking existing task adaptatio methods. We experiment with existing feature adaptation methods on a fair setting, which turn out to be ineffective on tasks that are highly different. Then we impose simple task-specific weight modulation on model weights that are sensitive to specific tasks. Due to practicability concerns, we perform rank analysis and figure out a suitable low-rank decomposition strategy for the task-specific bias on weights. Further beyond, we propose synchronous training that enables the model to learn general knowledge in the backbone weights and inject task-specific knowledge to the biases in an automatic fashion. We introduce the method to IPT and develop Instruct-IPT, which is the SOTA All-in-One image restoration model. The proposed model could achieve supreme performance on various restoration tasks, including denoising, deblurring, deraining, dehazing, and desnowing. Further, we introduce text commands as a good user interface. Beyond conventional regression-based low-level vision tasks, we also extended our All-in-One method to diffusion models to demonstrate the generalizability of our method.


\newpage
{
    \small
    \bibliographystyle{ieeenat_fullname}
    \bibliography{main}
}

\input{sec/X_suppl}

\end{document}

%% file: sec/X_suppl.tex
\clearpage
\appendix
\setcounter{page}{1}
\maketitlesupplementary

\section{Additional Experiments}
\textbf{General Knowledge Matters in Weight Modulation.} Via experiments, we are aware that certain types of weights are highly sensitive for different tasks. Apart from weight modulation that we introduce in the main body of this paper, we are also interested in direct weight replacement as a more radical form of weight modulation. Weight replacement (or "Mixture of Experts", MoE, in some works~\cite{qin2023unicontrol}) involves using different weights for different tasks. As output layers appear to be highly different (as shown in Tab.~\ref{tab:tasksim}) for different tasks, we conduct an experiment by imposing weight replacement on the last output layer.

\begin{table*}[htbp]
	\centering
	\setlength{\belowcaptionskip}{0cm}   
	\begin{tabular}{lcccc}
		\toprule
		& \multicolumn{2}{c}{Denoising} & \multicolumn{2}{c}{Deblurring} \\
		Methods & BSD & Urban100 & GoPro & HIDE \\
		\midrule
		Plain Mixed Training & \textbf{34.37} & \textbf{35.11} & \textbf{32.81} & \textbf{30.75} \\
		Weight Replacement (Output Layer) & 34.34 & 35.02 & 32.42 & 30.42 \\
		\bottomrule
	\end{tabular}
	\vspace{3pt}
	\caption{\textbf{Comparing Weight Replacement with Plain Mixed Training.} Weight Replacement of the last output layer hurts the performance of IPT when trained with two tasks together.}
	\label{tab:appendix_weight_replacement}
\end{table*}

\begin{table*}[htbp]
	\centering
	\setlength{\belowcaptionskip}{0cm}   
	\begin{tabular}{lccc}
		\toprule
		Methods & Params (\%) & GoPro & HIDE \\
		\midrule
		Plain Mixed Training & 0 & 32.81 & 30.75 \\
		\midrule
		Proportional Rank (Output Layer) & 29.95 & 33.30 & 31.16 \\
		Constant Rank (Output Layer) & \textbf{22.43} & \textbf{33.33} & \textbf{31.18} \\
		\bottomrule
	\end{tabular}
	\vspace{3pt}
	\caption{\textbf{Comparing different rank selection strategies in practice.} Constant rank are more effective: they perform better with fewer parameters.}
	\label{tab:appendix_rank}
\end{table*}

As shown in Tab.~\ref{tab:appendix_weight_replacement}, the performance of IPT decays due to the absence of the backbone weight. According to the finding that using task-specific weights could only hurt the performance, we conclude it is necessary to maintain a backbone weight for image restoration knowledge in general.

\noindent\textbf{The Practical Effectiveness of Constants over Proportional Rank.} In Sec.~\ref{sec:efficient_weight_adaptation}, we have demonstrated the rationality of the constant rank strategy via rank analysis. But is the constant rank strategy still more effective than proportional rank in practice? We accordingly design two bias config in addition to the IPT U-Net backbone and finetune them on the task of deblurring to verify the conclusion from rank analysis. Results in Tab.~\ref{tab:appendix_rank} reveals that the constant rank strategy is better in practice. This confirms the outcomes of our rank analysis.

\noindent\textbf{More Comparison with Single-Task Experts.} Here in the appendix, we provide more comparison with single-task experts on denoising (Tab.~\ref{tab:denoise}) and deblurring (Tab.~\ref{tab:deblur}). Our Instruct-IPT is in at a close margin with State-of-the-Art methods.

\begin{table*}[htbp]	\centering	\setlength{\belowcaptionskip}{0cm}   		\begin{tabular}{c| c c c c c  | c}		\toprule		&SwinIR&Restormer&GRL-B&ART&IPT-V2& Instruct-IPT\\		&~\cite{liang2021swinir}&~\cite{zamir2022restormer}&~\cite{grl}&~\cite{zhang2023accurate}&~\cite{iptv2}& (Ours) \\		\toprule		$\sigma=15$  &	34.42 & 34.40 &	 \underline{34.45} &	\textbf{34.46} & \textbf{34.46} & 34.40\\		$\sigma=25$  &	31.78 & 31.79 &	 \underline{31.82} &	\textbf{31.84} & \textbf{31.84} & 31.79\\		$\sigma=50$ &	28.56 & 28.60 &	 28.62 &	 \underline{28.63} & \textbf{28.65} & 28.61\\		\bottomrule	\end{tabular}	\vspace{3pt}	\caption{\textbf{Comparing Instruct-IPT with denoising experts.} PSNR on BSD68~\cite{martin2001database} is reported. The best and second best results are	\textbf{bolded} and \underline{underlined}.}	\label{tab:denoise}\end{table*}

\begin{table*}[!t]
	\centering
	\setlength{\belowcaptionskip}{0cm}   
	\scalebox{0.95}{
		\begin{tabular}{c c c c c c c c | c}
			\toprule
			Suin \textit{etc} &Cho \textit{etc} & IPT & MPRNet  & Restormer & NAF-Net &  DiffIR & GRL-B & Instruct-IPT\\
			\cite{suin2020spatially}   &\cite{cho2021rethinking}  &  \cite{chen2021pre} & \cite{zamir2021multi} & \cite{zamir2022restormer} & \cite{chen2022simple}  & \cite{xia2023diffir} & \cite{grl} & (Ours) \\
			\toprule
			31.85 &	32.45 &	32.52 &	32.66 &	32.92 & 33.71 &	 33.20 & \textbf{33.93} &  \underline{33.86}\\
			
			\bottomrule
	\end{tabular}}
	\vspace{3pt}
	\caption{\textbf{Comparing Instruct-IPT with deblurring experts.} PSNR on GoPro~\cite{nah2017deep} is reported. The best and second best results are \textbf{bolded} and \underline{underlined}.}
	\label{tab:deblur}
\end{table*}

\begin{table*}[!b]
    \centering
    \setlength{\belowcaptionskip}{0cm}  
    \begin{tabular}{l|cccc|cccc}
        \toprule
        & \multicolumn{4}{c|}{Inpainting} & \multicolumn{4}{c}{Outpainting} \\
        Method & LPIPS$\downarrow$ & DISTS$\downarrow$ & FID$\downarrow$ & IS$\uparrow$ & LPIPS$\downarrow$ & DISTS$\downarrow$ & FID$\downarrow$ & IS$\uparrow$ \\
        \midrule
        UniControl~\cite{qin2023unicontrol} & 0.2131 & 0.1193 & 10.4099 & 31.0591 & 0.3892 & 0.1813 & 12.4245 & 31.3619 \\
        Plain Mixed Ft. & 0.2143 & 0.1206 & 10.1269 & 31.5059 & 0.3860 & 0.1803 & 11.8556 & 31.8703 \\
        Bias only Ft. & 0.2122 & 0.1188 & 10.2626 & 31.5458 & 0.3856 & 0.1801 & 12.0940 & 31.1458 \\
        Sync. Ft.(Ours) & \textbf{0.2114} & \textbf{0.1183} & \textbf{10.1052} & \textbf{31.6833} & \textbf{0.3831} & \textbf{0.1784} & \textbf{11.8497} & \textbf{31.9936} \\
        \bottomrule
    \end{tabular}
    \vspace{-5pt}
    \caption{\textbf{Performance comparison of different methods on inpainting and outpainting.} The performance of our method is evaluated on COCO-Stuff~\cite{caesar2018coco}. A series of metrics are used to verify the effectiveness of our method over mixed finetuning and bias-only finetuning.}
    \label{tab:performance_comparison}
  \end{table*}




\noindent\textbf{Extrapolation to Diffusion.} Due to page limits, we record our contribution of the methods' extrapolation to diffusion models in this appendix section. We evaluate the quality of generated images using four metrics: LPIPS(Learned Perceptual Image Patch Similarity)~\cite{zhang2018perceptual}, DISTS(Deep Image Structure and Texture Similarity)~\cite{Ding_2020}, FID(Fréchet Inception Distance)~\cite{heusel2018gans}, and IS(Inception Score)~\cite{salimans2016improved}. These metrics assess different aspects of image similarity and quality, offering a comprehensive assessment of image generation quality, and highlighting the perceptual fidelity, structure consistency, distributional similarity, and diversity of generated images. As shown in Tab.~\ref{tab:performance_comparison}, we compared the performance of our proposed method against several baseline approaches: inference with original weights, direct fine-tuning, and fine-tuning only task-specific weight biases. Our results demonstrate that our method consistently outperforms these baselines across all evaluated metrics. We also present a visual qualitative comparison here. Our method could also outcompete baselines in terms of visual quality.

\begin{figure*}[htbp]
    \centering
    \includegraphics[width=\textwidth]{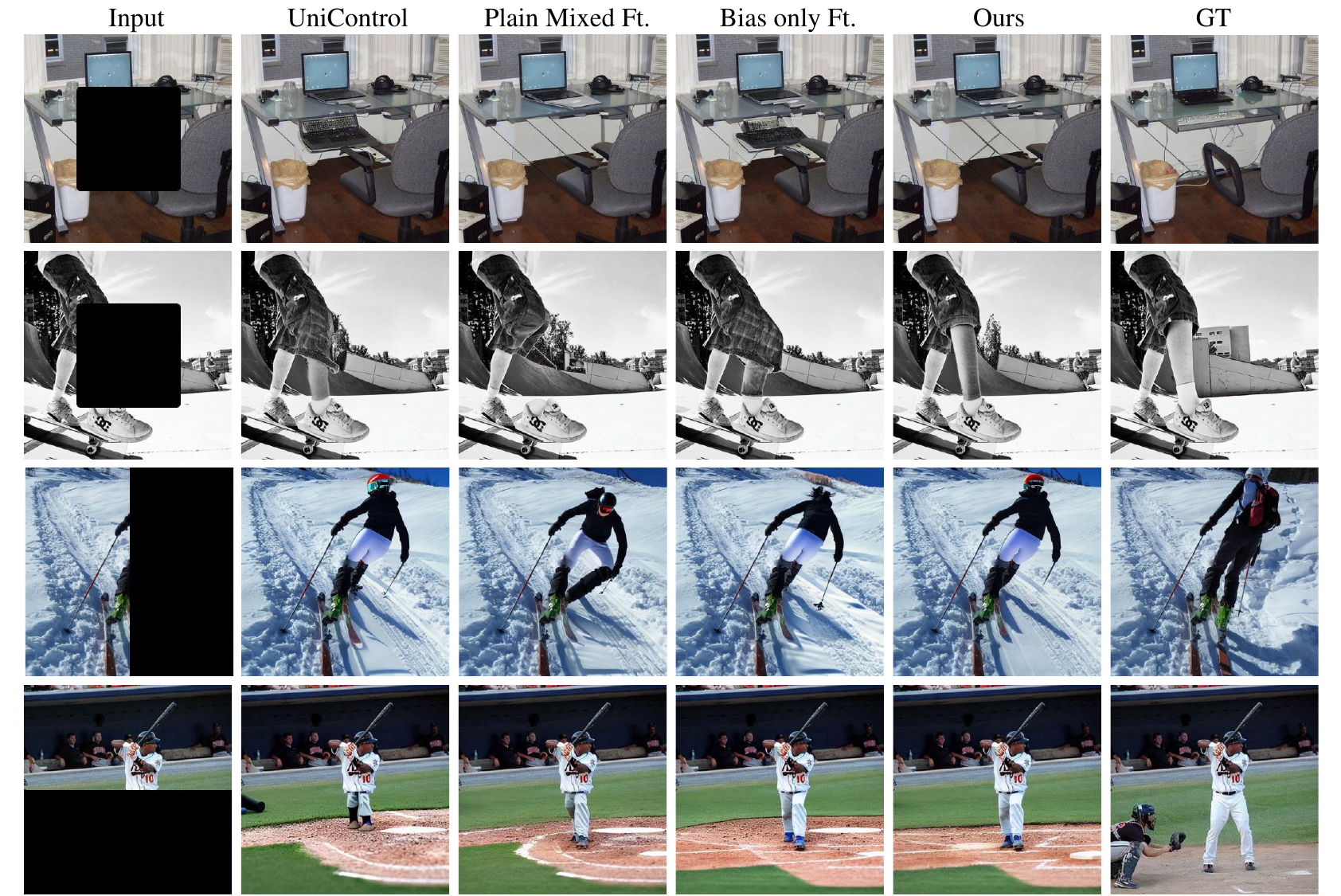}
    \caption{\textbf{Qualitative comparisons of several methods for diffusion models.} We compare our method with three other baselines on generative tasks: inpainting (the first two rows) and outpainting (the last two rows). Our method generates images with greater logical consistency and realism. }
    \label{fig:diffusion}
\end{figure*}

Fig.~\ref{fig:diffusion} presents a qualitative comparison, showcasing our method's ability to produce more visually compelling and appealing results compared to other baselines.

